\documentclass[conference]{IEEEtran}
\IEEEoverridecommandlockouts
\usepackage{cite}
\usepackage{amsmath,amssymb,amsfonts}
\usepackage{algorithmic}
\usepackage{graphicx}
\usepackage{textcomp}
\usepackage{xcolor}
\usepackage{url}
\usepackage{multirow}
\def\BibTeX{{\rm B\kern-.05em{\sc i\kern-.025em b}\kern-.08em
    T\kern-.1667em\lower.7ex\hbox{E}\kern-.125emX}}
\begin{document}

\title{LLM-as-a-Judge for Voice-Agent Evaluation: Alignment, Stability, and Human Oversight}

\author{
\IEEEauthorblockN{Shashank Singh\textsuperscript{\textdagger}, Anupam Purwar\textsuperscript{\textdagger,*}, Kritika Srivastava }
\IEEEauthorblockA{\textit{Sprinklr AI}\\
India}
}

\author{
 Shashank Singh\textsuperscript{\textdagger} \\
 Sprinklr AI \\
 Bengaluru, India
 \and
 Anupam Purwar\textsuperscript{\textdagger,*} \\
 Sprinklr AI\\
 Gurugram, India
 \thanks{\textdagger~These authors contributed equally to this work.}
 \thanks{*Corresponding author: \texttt{anupam.aiml@gmail.com}. Project page: \texttt{https://anupam-purwar.github.io/page/}}
 \and
 Kritika Srivastava \\
 Sprinklr AI \\
 Bengaluru, India
}

\maketitle

\begin{abstract}
Evaluating conversational voice agents at scale requires reliable assessment methods that capture both observable interaction quality and the contextual judgment typically provided by human evaluators. We investigate LLM-as-a-Judge evaluation by comparing human judgments with GPT-4.1 and GPT-5 on telecom and retail voice-agent conversations, across conversational quality and safety dimensions. The same interactions are scored under three evaluation configurations, p0, p1, and p2, to test whether automated judgments are sensitive to the evaluation setup and whether observed patterns generalize across configurations and judge models. Beyond aggregate agreement, we examine metric-level correlations, evaluator consistency, and systematic human-LLM disagreement to identify which conversational attributes can be judged reliably by automation and which remain sensitive to interpretation and context. Effective voice-agent evaluation is also shaped by pipeline-level factors such as speech generation, streaming, and error propagation across ASR, reasoning, and tool-calling stages, motivating our focus on comparing how human and LLM judges score the same interactions end to end. Our results show that LLM-based evaluation can serve as an effective component of large-scale voice-agent assessment, but that its reliability is metric- and configuration-dependent rather than uniform. This provides an empirical framework for identifying which metrics suit automated evaluation and supports hybrid pipelines in which LLM judges handle scalable assessment while human evaluators remain engaged for metrics that demand contextual interpretation and higher-confidence judgment.
\end{abstract}

\section{Introduction}
Evaluating conversational voice agents has traditionally relied on human evaluation, which, while providing a reliable measure of interaction quality, is expensive, time-consuming, subjective, and difficult to scale across large volumes of conversations \cite{b1} \cite{b2} . With the recent advances in Large Language Models (LLMs), automated evaluation using LLMs as judges has emerged as a promising alternative for assessing open-ended conversations across multiple quality dimensions \cite{b3} \cite{b4}. Prior work on dialogue-system evaluation has explored both human and automated evaluation protocols, highlighting challenges in obtaining reliable and consistent human judgments as well as the limitations of conventional automatic metrics in capturing nuanced conversational quality \cite{b1} \cite{b2} \cite{b5}. More recently, the LLM-as-a-Judge paradigm has demonstrated that capable LLMs can achieve substantial agreement with human preferences when evaluating multi-turn conversations, establishing the potential of foundation models as scalable evaluators \cite{b3}. However, subsequent studies have shown that LLM-based judges can exhibit position bias, prompt sensitivity, and variability across evaluation settings, raising questions about the stability and reliability of automated judgments \cite{b6} \cite{b7}. Surveys of LLM-based agent evaluation further indicate that conversational quality is inherently multi-dimensional, encompassing aspects such as response quality, task completion, contextual understanding, consistency, and user experience, and that different evaluation dimensions may exhibit different levels of agreement between LLMs and human evaluators \cite{b7} \cite{b8}. Voice agents introduce additional layers of complexity that text-only dialogue evaluation does not have to contend with. Prior work on low-latency Voice-to-Voice architectures shows that end-to-end responsiveness in such systems depends not just on the underlying language model but also on speech generation and streaming \cite{b18}. Separately, work on benchmarking multi-modal agents has argued that errors can originate and compound at any stage of the pipeline, from speech recognition through reasoning, tool use, and speech synthesis \cite{b19}. Taken together, these findings suggest that a full picture of voice-agent quality cannot come from evaluating the conversation transcript alone, and this is part of what motivates our focus on comparing how human and LLM judges score the same interactions. These findings motivate a systematic investigation into the extent to which LLM-based evaluation can serve as a reliable complement to human assessment for conversational voice agents. In this work, we compare human evaluations with LLM-as-a-Judge evaluations using GPT-4.1 and GPT-5 across three evaluation configurations, namely p0, p1, and p2. We investigate (1) whether LLM evaluations remain stable across evaluation modes, (2) how closely LLM assessments align with human judgments, and (3) which evaluation metrics can be reliably automated and which continue to require human oversight.

To make the evaluation conditions explicit, in this work, we compare human evaluations with LLM-as-a-Judge evaluations using GPT-4.1 and GPT-5 across three evaluation configurations (namely p0,p1,p2)\cite{b9}:

\begin{itemize}
    \item p0 (no user context) - No Persona: No additional persona or user-context information is provided to the agent. The agent responds based solely on the ongoing conversation.
    \item p1 (static persona) - Persona Injection: A predefined user persona is explicitly provided to the agent, including characteristics such as domain expertise, ambiguity, and communication behavior.
    \item p2 (dynamically inferred context) - Context Injection: The agent receives dynamically inferred user context derived from the ongoing conversation. This enables the agent to adapt to changes in the user’s behavior, expertise, or interaction style during the conversation.
\end{itemize}

Through this comparison, we aim to characterize the reliability, consistency, and practical applicability of LLM-based evaluation for conversational voice agents and identify the evaluation dimensions for which automated judging can effectively complement human assessment.
\begin{figure*}[tbp]
    \centering
    \includegraphics[width=\textwidth]{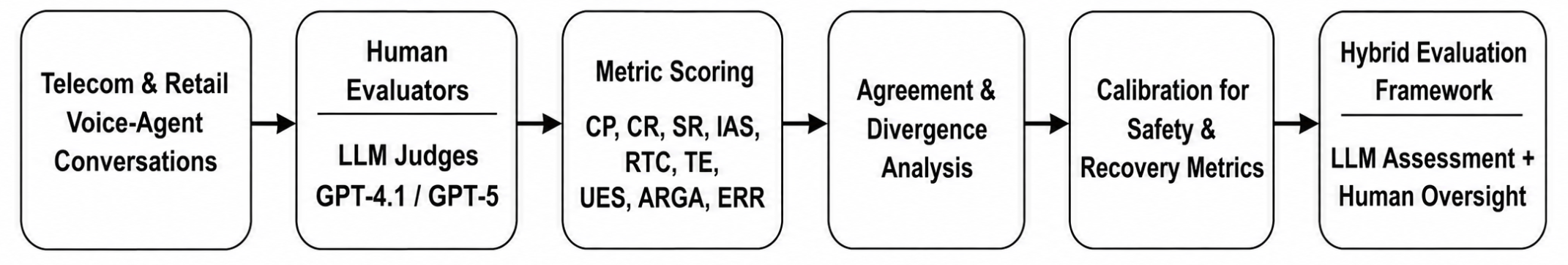}
    \caption{Workflow for comparing human and LLM-as-Judge evaluations, from voice-agent conversation scoring through divergence analysis, calibration, and hybrid evaluation design.}
    \label{fig:pipeline}
\end{figure*}

\begin{table*}[htbp]
\caption{Task-Type Distribution Across Retail and Telecom Domains}
\label{tab:task_distribution}
\centering
\renewcommand{\arraystretch}{1.3}
\begin{tabular}{|l|l|c|p{7cm}|}
\hline
\textbf{Domain} & \textbf{Configuration Counts} & \textbf{Domain Total} & \textbf{Task-Type Distribution} \\
\hline
\multirow{6}{*}{Retail} & \multirow{6}{*}{P0: 40, P1: 40, P2: 40} & \multirow{6}{*}{120} & Return / Exchange / Modification / Cancellation: 98 \\
 & & & Address Update / Change: 11 \\
 & & & Order Information / Inquiry: 8 \\
 & & & Tracking Request: 1 \\
 & & & Payment Method Change: 1 \\
 & & & Account Verification / Human Transfer: 1 \\
\hline
\multirow{3}{*}{Telecom} & \multirow{3}{*}{P0: 41, P1: 40, P2: 41} & \multirow{3}{*}{122} & MMS / Picture Messaging Issue: 56 \\
 & & & No Service / Service Restoration: 42 \\
 & & & Mobile Data / Internet Connectivity Issue: 24 \\
\hline
Overall & 6 configurations & 242 & Retail: 120; Telecom: 122 \\
\hline
\end{tabular}
\end{table*}

\begin{table*}[htbp]
\caption{Ratio of Human to LLM-Judge Scores Across Evaluation Modes (Telecom)}
\label{tab:human_llm_ratio_telecom}
\centering
\scriptsize
\setlength{\tabcolsep}{4pt}
\renewcommand{\arraystretch}{1.2}
\begin{tabular}{|l|c|c|c|c|c|c|}
\hline
\textbf{Metric Ratios} & \textbf{H/G4.1 (p0)} & \textbf{H/G5 (p0)} & \textbf{H/G4.1 (p1)} & \textbf{H/G5 (p1)} & \textbf{H/G4.1 (p2)} & \textbf{H/G5 (p2)} \\
\hline
CFA (Critical Field Accuracy) & 1.717 & 1.684 & 1.716 & 1.709 & 1.814 & 1.799 \\
\hline
CP (Confirmation Precision) & 2.200 & 2.138 & 4.040 & 4.301 & 2.381 & 2.738 \\
\hline
CR (Confirmation Recall) & 1.041 & 1.016 & 1.125 & 1.000 & 1.059 & 1.000 \\
\hline
IAS (Irreversible Action Safety) & 5.497 & 6.067 & 5.865 & 6.082 & 4.033 & 4.284 \\
\hline
SR (Safety Recall) & 5.644 & 5.323 & 5.133 & 4.852 & 3.479 & 3.071 \\
\hline
ARGA (ASR-Robust Goal Achievement) & 0.171 & 0.156 & 0.265 & 0.215 & 0.142 & 0.117 \\
\hline
TE (Turn Efficiency) & 0.845 & 0.827 & 0.966 & 0.945 & 0.913 & 0.893 \\
\hline
UES (User Experience Score) & 0.484 & 0.503 & 0.555 & 0.557 & 0.467 & 0.483 \\
\hline
ERR (Error Rate) & 1.120 & 1.733 & 1.461 & 2.705 & 0.857 & 2.020 \\
\hline
RTC (Recovery Turn Count) & 3.864 & 1.889 & 6.064 & 2.051 & 3.200 & 1.505 \\
\hline
\end{tabular}
\vspace{2pt}
\begin{flushleft}
\footnotesize \textit{Note: H/G4.1= Human score/LLM Score, where LLM = GPT 4.1 and H/G5= Human score/LLM Score, where LLM = GPT 5, respectively. p0, p1, p2 indicate evaluation configuration modes.  A ratio greater than 1 indicates that the human score exceeds the corresponding LLM-judge score, while a ratio below 1 indicates that the LLM judge assigns a higher score.}

\end{flushleft}
\end{table*}

\begin{table*}[htbp]
\caption{Ratio of Human to LLM-Judge Scores Across Evaluation Modes (Retail)}
\label{tab:human_llm_ratio_retail}
\centering
\scriptsize
\setlength{\tabcolsep}{4pt}
\renewcommand{\arraystretch}{1.2}
\begin{tabular}{|l|c|c|c|c|c|c|}
\hline
\textbf{Metric Ratios} & \textbf{H/G4.1 (p0)} & \textbf{H/G5 (p0)} & \textbf{H/G4.1 (p1)} & \textbf{H/G5 (p1)} & \textbf{H/G4.1 (p2)} & \textbf{H/G5 (p2)} \\
\hline
CFA (Critical Field Accuracy) & 1.394 & 1.204 & 1.142 & 0.986 & 1.128 & 1.023 \\
\hline
CP (Confirmation Precision) & 0.659 & 0.953 & 0.567 & 0.924 & 0.492 & 1.934 \\
\hline
CR (Confirmation Recall) & 0.730 & 0.738 & 0.909 & 0.837 & 0.962 & 0.962 \\
\hline
IAS (Irreversible Action Safety) & 1.418 & 1.754 & 1.473 & 1.623 & 1.189 & 1.441 \\
\hline
SR (Safety Recall) & 1.292 & 1.543 & 1.427 & 1.626 & 1.056 & 1.269 \\
\hline
ARGA (ASR-Robust Goal Achievement) & 0.562 & 0.598 & 0.448 & 0.657 & 0.387 & 0.383 \\
\hline
TE (Turn Efficiency) & 0.847 & 0.847 & 0.763 & 0.755 & 0.900 & 0.900 \\
\hline
UES (User Experience Score) & 0.464 & 0.496 & 0.482 & 0.527 & 0.440 & 0.466 \\
\hline
ERR (Error Rate) & 3.855 & 2.998 & 5.665 & 5.665 & 6.024 & 2.906 \\
\hline
RTC (Recovery Turn Count) & 1.556 & 0.791 & 1.469 & 1.063 & 1.148 & 2.270 \\
\hline
\end{tabular}
\begin{flushleft}
\footnotesize \textit{Note: H/G4.1= Human score/LLM Score, where LLM = GPT 4.1 and H/G5= Human score/LLM Score, where LLM = GPT 5, respectively. p0, p1, p2 indicate evaluation configuration modes.  A ratio greater than 1 indicates that the human score exceeds the corresponding LLM-judge score, while a ratio below 1 indicates that the LLM judge assigns a higher score.}
\end{flushleft}
\end{table*}
\vspace{2pt}

\section{Evaluation Methodology}\label{sec:metrics}
Figure~\ref{fig:pipeline} summarizes the overall workflow used in this study, from scoring voice-agent conversations through divergence analysis, calibration, and the design of a hybrid human-LLM evaluation pipeline. Human evaluators and LLM judges independently assessed the same sets of retail and telecom voice-agent conversations across multiple metrics. Each conversation in the retail and telecom evaluation sets was independently scored by $n=3$ trained human annotators using the same rubric definitions supplied to the LLM judges. The human scores used in the human-LLM ratios (Tables~\ref{tab:human_llm_ratio_telecom} and~\ref{tab:human_llm_ratio_retail}) are those of Manual Eval~1, not the mean of the three annotators (Section~\ref{sec:manual_crosscheck}). Inter-annotator agreement was monitored during annotation, and conversations with high annotator disagreement were flagged for adjudication by a fourth senior annotator rather than being resolved by simple majority vote. The same aggregation procedure (mean across the conversation set) is applied uniformly to both the Eval~1 human scores and the LLM-judge scores so that the human-to-LLM ratios reported in Table~\ref{tab:human_llm_ratio_telecom} are directly comparable.
The metrics evaluated by both human annotators and LLM judges, defined and reported in benchmark paper MM-$\tau$-p\textsuperscript{2}~\cite{b9} (refer Appendix~\ref{app:metric_defs}).
These metric definitions and their rubric-based scoring criteria are adopted unchanged from our companion benchmark paper, MM-$\tau$-p\textsuperscript{2}~\cite{b9}, which introduces them alongside seven additional metrics spanning goal achievement, efficiency, recovery, and safety for multi-modal agent evaluation. For each metric, the ratio between human and LLM-generated scores was calculated for GPT-4.1 and GPT-5 under p0, p1, and p2 evaluation modes. This comparison enabled a quantitative assessment of evaluator agreement, highlighting systematic differences in scoring behavior across models and prompting configurations. The analysis also examined the consistency of these differences across individual metrics, allowing us to identify evaluation dimensions where LLM judgments closely align with human assessments and those where notable deviations persist.

Across the two verticals, the evaluation set spanned 242 conversations in total, drawn from six separate configurations, three prompting modes (p0, p1, p2) applied to each of the Retail and Telecom domains. Retail accounted for 120 cases, split evenly at 40 per configuration, while Telecom contributed 122 cases (41, 40, and 41 across the three modes, refer Table~\ref{tab:task_distribution}). Within Retail, the vast majority of cases involved return, exchange, modification, or cancellation requests (98 of 120), with the remainder covering address changes, order information inquiries, tracking requests, payment method changes, and a small number of account verification or human-transfer cases. Telecom conversations were more evenly distributed across three issue types: MMS or picture-messaging problems accounted for 56 cases, service outages or restoration requests for 42, and mobile data or connectivity issues for the remaining 24. This distribution reflects the kinds of problems each domain typically surfaces in real customer interactions, and it also means the evaluation results are naturally weighted toward the task types that occur most often, something worth keeping in mind when comparing metric behavior across domains.

\subsection{Evaluation Process for Voice Agent Conversations (Retail/Telecom)}

Each evaluation begins with opening the conversation log, either Retail or Telecom, and reading through the exchange in full before any scoring takes place. Every case includes three versions of the same interaction: the ground truth, representing how the conversation should ideally have unfolded, alongside the actual voice conversation and text conversation. Having all three side by side makes it possible to see precisely where the agent's real behavior deviated from the expected flow, and how its performance differed between speaking and typing. Voice interactions are generally more error-prone than text, so particular attention is paid to identifying ASR (speech recognition) errors and evaluating how well the agent recovers from them. While reviewing the transcript, unnecessary clarifications or confirmations are flagged, and necessary ones are marked separately so they are easy to locate later. All ASR errors are highlighted in red for quick visibility. Whenever an error occurs, the number of turns the agent takes to recover from it is counted. User effort is assessed alongside this, specifically how much the customer had to repeat, correct, or restate before the task was completed. Whether the intended goal was ultimately achieved is also checked in every case, and where a conversation was escalated to a human agent, the likely cause is investigated to understand what the AI agent was unable to resolve on its own. In terms of pace, a full working day - roughly six hours, from 12 PM to 6 PM - typically covers 10 to 15 cases, averaging around 24 to 36 minutes per case. This figure is approximate rather than fixed, since evaluation time depends on the length of the conversation and the number of errors that need to be traced through for recovery analysis; longer or more error-dense cases naturally take more time to review thoroughly.

\subsection{Multi-Annotator Manual Scores}
\label{sec:manual_crosscheck}

We collected scores from three independent human annotators on all twelve metrics for both domains. Tables~\ref{tab:manual_p0}, \ref{tab:manual_p1}, and \ref{tab:manual_p2} report the metric-level aggregates. Table~\ref{tab:manual_p0} covers configuration p0. Table~\ref{tab:manual_p1} covers p1. Table~\ref{tab:manual_p2} covers p2. Each table lists Manual Eval~1, Manual Eval~2, and Manual Eval~3 for Retail and Telecom. MRS is constant at $1.0$ in nearly all cells. TO is $0$ throughout. Neither metric adds any variance, so both are left out of the agreement analysis below.

Manual Eval~1 matches the human baselines used in Tables~\ref{tab:human_llm_ratio_telecom} and~\ref{tab:human_llm_ratio_retail}. Agreement here is measured on those aggregates, not on individual conversations, so it is not conversation-level inter-annotator agreement in the usual sense. For the nine bounded metrics in the main ratio tables, excluding RTC, pairwise Pearson correlation is $r = 0.975$ between Eval~1 and Eval~3 and $r = 0.903$ between Eval~1 and Eval~2 ($N = 54$ cells: 9 metrics, 2 domains, 3 configurations). Including RTC reduces these values to $r = 0.930$ and $r = 0.804$ ($N = 60$). Retail Eval~3 is identical across p0, p1, and p2, which inflates the Eval~1 to Eval~3 correlation in that domain.

To compare annotator spread across metrics on different scales, we use the mean relative range: for each of the six domain and configuration cells we take the range across the three annotators divided by their mean, then average over the six cells. On this measure the ordering is ARGA ($0.77$), RTC ($0.46$), CP ($0.27$), SR ($0.20$), ERR ($0.20$), IAS ($0.15$), UES ($0.15$), CR ($0.15$), TE ($0.14$), and CFA ($0.11$). ARGA, RTC, and CP are therefore the least reproducible metrics across annotators, and CFA, TE, CR, and UES the most reproducible. The ordering differs if absolute range is used instead, because ARGA and CP have small means: by absolute range, SR ($0.15$), CP ($0.15$), IAS ($0.13$), and CR ($0.13$) are comparable. We use relative range throughout because the metrics differ in scale. Annotator spread and human-LLM divergence are distinct: the largest human-LLM ratios are on IAS and SR, which are mid-ranked for annotator spread.

\begin{table*}[htbp]
\caption{Manual Evaluation Scores for Configuration p0 (No Persona)}
\label{tab:manual_p0}
\centering
\scriptsize
\setlength{\tabcolsep}{4pt}
\renewcommand{\arraystretch}{1.15}
\begin{tabular}{|l|ccc|ccc|}
\hline
\multirow{2}{*}{\textbf{Metric}} & \multicolumn{3}{c|}{\textbf{Retail}} & \multicolumn{3}{c|}{\textbf{Telecom}} \\
\cline{2-7}
 & \textbf{Eval~1} & \textbf{Eval~2} & \textbf{Eval~3} & \textbf{Eval~1} & \textbf{Eval~2} & \textbf{Eval~3} \\
\hline
CFA  & 0.530 & 0.522 & 0.525 & 0.815 & 0.769 & 0.784 \\
\hline
ARGA & 0.287 & 0.248 & 0.258 & 0.055 & 0.112 & 0.074 \\
\hline
MRS  & 1.000 & 1.000 & 1.000 & 1.000 & 1.000 & 1.000 \\
\hline
TE   & 0.831 & 0.844 & 0.852 & 0.817 & 0.745 & 0.793 \\
\hline
TO   & 0.000 & 0.000 & 0.000 & 0.000 & 0.000 & 0.000 \\
\hline
UES  & 0.213 & 0.205 & 0.203 & 0.215 & 0.185 & 0.205 \\
\hline
ERR  & 0.270 & 0.303 & 0.306 & 0.277 & 0.323 & 0.291 \\
\hline
RTC  & 1.385 & 2.500 & 2.381 & 2.300 & 2.797 & 2.741 \\
\hline
CP   & 0.448 & 0.456 & 0.462 & 0.364 & 0.394 & 0.385 \\
\hline
CR   & 0.716 & 0.679 & 0.743 & 1.000 & 1.059 & 1.059 \\
\hline
IAS  & 0.667 & 0.640 & 0.653 & 0.978 & 0.865 & 0.876 \\
\hline
SR   & 0.556 & 0.608 & 0.573 & 0.935 & 0.987 & 0.953 \\
\hline
\end{tabular}
\end{table*}

\begin{table*}[htbp]
\caption{Manual Evaluation Scores for Configuration p1 (Static Persona)}
\label{tab:manual_p1}
\centering
\scriptsize
\setlength{\tabcolsep}{4pt}
\renewcommand{\arraystretch}{1.15}
\begin{tabular}{|l|ccc|ccc|}
\hline
\multirow{2}{*}{\textbf{Metric}} & \multicolumn{3}{c|}{\textbf{Retail}} & \multicolumn{3}{c|}{\textbf{Telecom}} \\
\cline{2-7}
 & \textbf{Eval~1} & \textbf{Eval~2} & \textbf{Eval~3} & \textbf{Eval~1} & \textbf{Eval~2} & \textbf{Eval~3} \\
\hline
CFA  & 0.434 & 0.412 & 0.525 & 0.835 & 0.718 & 0.805 \\
\hline
ARGA & 0.197 & 0.652 & 0.258 & 0.061 & 0.273 & 0.134 \\
\hline
MRS  & 1.000 & 0.755 & 1.000 & 1.000 & 1.000 & 1.000 \\
\hline
TE   & 0.748 & 0.544 & 0.852 & 0.930 & 0.872 & 0.909 \\
\hline
TO   & 0.000 & 0.000 & 0.000 & 0.000 & 0.000 & 0.000 \\
\hline
UES  & 0.227 & 0.173 & 0.203 & 0.230 & 0.257 & 0.238 \\
\hline
ERR  & 0.283 & 0.333 & 0.306 & 0.267 & 0.364 & 0.278 \\
\hline
RTC  & 3.189 & 2.571 & 2.381 & 1.318 & 4.875 & 2.030 \\
\hline
CP   & 0.397 & 0.868 & 0.462 & 0.482 & 0.833 & 0.624 \\
\hline
CR   & 0.837 & 0.995 & 0.743 & 1.000 & 1.037 & 0.983 \\
\hline
IAS  & 0.795 & 0.818 & 0.653 & 1.000 & 0.800 & 0.904 \\
\hline
SR   & 0.699 & 0.857 & 0.573 & 0.828 & 1.043 & 0.779 \\
\hline
\end{tabular}
\end{table*}

\begin{table*}[htbp]
\caption{Manual Evaluation Scores for Configuration p2 (Dynamically Inferred Context)}
\label{tab:manual_p2}
\centering
\scriptsize
\setlength{\tabcolsep}{4pt}
\renewcommand{\arraystretch}{1.15}
\begin{tabular}{|l|ccc|ccc|}
\hline
\multirow{2}{*}{\textbf{Metric}} & \multicolumn{3}{c|}{\textbf{Retail}} & \multicolumn{3}{c|}{\textbf{Telecom}} \\
\cline{2-7}
 & \textbf{Eval~1} & \textbf{Eval~2} & \textbf{Eval~3} & \textbf{Eval~1} & \textbf{Eval~2} & \textbf{Eval~3} \\
\hline
CFA  & 0.439 & 0.437 & 0.525 & 0.866 & 0.866 & 0.868 \\
\hline
ARGA & 0.225 & 0.342 & 0.258 & 0.037 & 0.085 & 0.067 \\
\hline
MRS  & 1.000 & 1.000 & 1.000 & 1.000 & 1.000 & 1.000 \\
\hline
TE   & 0.887 & 0.920 & 0.852 & 0.874 & 0.756 & 0.804 \\
\hline
TO   & 0.000 & 0.000 & 0.000 & 0.000 & 0.000 & 0.000 \\
\hline
UES  & 0.223 & 0.193 & 0.203 & 0.210 & 0.173 & 0.193 \\
\hline
ERR  & 0.247 & 0.228 & 0.306 & 0.309 & 0.267 & 0.275 \\
\hline
RTC  & 2.951 & 3.618 & 2.381 & 1.379 & 1.396 & 1.451 \\
\hline
CP   & 0.478 & 0.453 & 0.462 & 0.397 & 0.419 & 0.393 \\
\hline
CR   & 0.962 & 0.846 & 0.743 & 1.000 & 1.130 & 1.128 \\
\hline
IAS  & 0.620 & 0.667 & 0.653 & 1.000 & 0.794 & 0.826 \\
\hline
SR   & 0.518 & 0.436 & 0.573 & 0.821 & 0.906 & 0.871 \\
\hline
\end{tabular}
\end{table*}

\section{Results}
\subsection{Analysis for Telecom sector conversations}
Table~\ref{tab:human_llm_ratio_telecom} splits the Telecom metrics into two groups. In the first group the two evaluator types stay close. Turn Efficiency stays near a ratio of 1 in all three configurations and for both judges. Confirmation Recall stays near 1.0 to 1.1. Humans and judges agree on turn efficiency and on when a clarification was needed. User Experience Score is not in this group. Its ratios are near 0.5. The judges give a higher UES than humans do.

The second group diverges. Irreversible Action Safety and Safety Recall have ratios above 3 in every configuration. Humans record far more safety events than either judge. Confirmation Precision reaches 4.3 under p1. There the judges penalize unnecessary clarifications more heavily than humans. ARGA moves the other way. Its ratio stays below 1, from 0.117 to 0.265. The judges are more generous than humans on goal achievement after an ASR error. Recovery Turn Count is the most volatile metric in the table. It ranges from 1.889 to 6.064.

GPT-4.1 and GPT-5 track each other closely on most metrics. GPT-5 is slightly closer to human scores on CFA and TE. GPT-4.1 is slightly closer on IAS and SR. Neither model leads across the full metric set.

Tables~\ref{tab:manual_p0}, \ref{tab:manual_p1}, and \ref{tab:manual_p2} show the same Telecom conversations scored by three annotators. Manual Eval~1 is the baseline used in Table~\ref{tab:human_llm_ratio_telecom}. CFA, TE, UES, ERR, and CR are more reproducible across annotators than RTC or CP. They are still not uniformly tight. CFA spreads under p1, at 0.835, 0.718, and 0.805. TE spreads under p2, at 0.874, 0.756, and 0.804. UES is the tightest of the five. Its annotator range is at most 0.037 in any Telecom configuration.

The safety scores are high for every annotator. IAS runs from 0.794 to 1.000 and SR from 0.779 to 1.043. The spread is wider under p1. The safety gap against the judges therefore holds for all three annotators, not only Eval~1. ARGA is low for all three, from 0.037 to 0.273. This agrees with the ratio result. Humans score ASR-robust goal achievement more strictly than the judges do. The size of the gap depends on the annotator under p1, where Eval~2 reports 0.273 against 0.061 and 0.134.

Annotator disagreement is largest on RTC and CP under p1. Telecom RTC in p1 is 1.318, 4.875, and 2.030. CP is 0.482, 0.833, and 0.624. p0 and p2 are much tighter on RTC, with ranges of 0.497 and 0.072. The volatility seen for RTC in Table~\ref{tab:human_llm_ratio_telecom} is therefore not only a judge problem. Humans also disagree on how many recovery turns to count. The study has one persona-injection condition per domain. On this evidence the p1 spread cannot be assigned to persona injection by itself.
\subsection{Analysis for Retail sector conversations}
The Retail metrics split the same way (Table~\ref{tab:human_llm_ratio_retail}). Turn Efficiency and Confirmation Recall stay near a ratio of 1 in all three configurations. Critical Field Accuracy is also fairly stable, from 0.986 to 1.394. Both evaluator types score field-level correctness in much the same way.

The safety metrics again show the largest gaps. The gaps are smaller than in Telecom. IAS runs from 1.189 to 1.754 and SR from 1.056 to 1.626. Both stay above 1. Confirmation Precision is the least stable metric in the table. Under p2 it moves from 0.492 for GPT-4.1 to 1.934 for GPT-5. The two judges therefore disagree with each other, not only with the human baseline. Error Rate is also erratic, from 2.906 to 6.024. ARGA stays below 1, from 0.383 to 0.657. The judges again score ASR-error recovery more generously than humans. Recovery Turn Count has a moderate spread, from 0.791 to 2.270.

Humans apply stricter safety judgment in Retail. The judges are more lenient on ASR-robust goal achievement. Confirmation Precision and Error Rate are the least reliable metrics for automated judging in this domain.

The annotators agree most closely under p0 (Tables~\ref{tab:manual_p0}, \ref{tab:manual_p1}, and \ref{tab:manual_p2}). For Retail p0 they are within a narrow band on CFA (0.522 to 0.530), TE (0.831 to 0.852), UES (0.203 to 0.213), and CP (0.448 to 0.462). IAS and SR also cluster, at 0.640 to 0.667 and 0.556 to 0.608. These safety scores are lower than the Telecom values in the same configuration. That is consistent with the smaller Retail safety ratios.

Configuration p1 has the widest annotator spread. Eval~2 reports ARGA of 0.652 against 0.197 and 0.258 for the other two. It reports CP of 0.868 against 0.397 and 0.462. Its TE is also lower, at 0.544 against 0.748 and 0.852. Retail RTC varies in every configuration. It is 1.385, 2.500, and 2.381 under p0, and 2.951, 3.618, and 2.381 under p2. Recovery and clarification precision are the hardest metrics for humans to score consistently.

One limit applies to Retail only. Eval~3 is unchanged across p0, p1, and p2 on every metric. It therefore gives no independent check at the configuration level. Comparisons across configurations in this domain rest on Eval~1 and Eval~2.
\subsection{Telecom vs. Retail}
The two domains disagree in the same direction but not by the same amount. In both, Turn Efficiency and Confirmation Recall stay near a ratio of 1 and ARGA stays below 1. Stable turn-level scoring and lenient judge scoring on ASR-error recovery are therefore not specific to one domain.

The safety metrics diverge much more in Telecom. IAS and SR ratios exceed 4 in Telecom in every configuration. In Retail they stay between 1.06 and 1.75. CFA is tighter in Retail, from 0.986 to 1.394, against 1.684 to 1.814 in Telecom. Confirmation Precision behaves differently in the two domains. In Telecom it rises to 4.3 under p1, where the judges over-penalize unnecessary clarifications. In Retail it falls to 0.492 and then rises to 1.934 for GPT-5 under p2. That swing reflects disagreement between the two judges rather than a steady bias against humans. RTC is more volatile in Telecom, from 1.889 to 6.064, than in Retail, from 0.791 to 2.270.

The annotator tables support part of this contrast. In Telecom all three annotators give high IAS and SR in every configuration. Retail values are generally lower, from 0.620 to 0.818 for IAS and 0.436 to 0.857 for SR. The two ranges do overlap in one cell. Under p1, Eval~2 gives a higher IAS in Retail (0.818) than in Telecom (0.800). SR is lower in Retail for every annotator and configuration.

The human baseline alone does not explain the smaller Retail safety ratios. Human IAS and SR do differ by domain. That difference is smaller than the difference in the ratios. The judges must therefore also be closer to humans on Retail safety than on Telecom safety.

Annotator spread and human-judge divergence do not fall on the same metrics. ARGA, RTC, and CP have the widest annotator spread in both domains. The widest human-judge gaps are on IAS and SR. The sharpest human disagreement is Telecom RTC under p1, where Eval~2 gives 4.875 against 1.318 and 2.030. Retail p1 shows the same pattern for ARGA and CP, again from Eval~2. Reading the tables qualitatively, using Eval~2 or Eval~3 in place of Eval~1 would not reverse the direction of the CFA, TE, CR, IAS, and SR ratios. It would change the size of the RTC, ARGA, and CP ratios, because the human score itself moves on those. We did not recompute the ratios under the other two annotators.

In summary, the direction of disagreement is the same in both domains. Humans are stricter on safety. The judges are more generous on ASR-robust recovery. The disagreement is larger in Telecom. Automated judging tracks human scores more closely in Retail. RTC, ARGA, and CP stay difficult for humans and judges alike.

\section{Discussion}

\subsection{Stability Across Evaluation Modes}\label{sec:correlation}
Most metrics keep the same relative pattern across p0, p1, and p2, and across GPT-4.1 and GPT-5. Humans and judges often differ in absolute score. The ranking of metrics stays much the same. Metrics that score high in one configuration still score high in the others. Metrics that score low stay low. This looks like a calibration difference rather than a change in what each evaluator treats as strong or weak.

The annotator tables qualify this on the human side, and the qualification depends on the domain. In Telecom the Eval~1 baseline is stable across modes on most metrics. CFA moves by $0.051$ across the three configurations. CR does not move at all. IAS moves by $0.022$, ARGA by $0.023$, and UES by $0.020$. In Retail the same baseline moves much more. CR runs from $0.716$ to $0.962$. SR runs from $0.518$ to $0.699$. IAS runs from $0.620$ to $0.795$. TE runs from $0.748$ to $0.887$. RTC moves in both domains, by $0.982$ turns in Telecom and $1.804$ turns in Retail. Mode insensitivity therefore holds well for the Telecom human baseline and for the judges. It holds less well for the Retail human baseline.

Two further limits apply. Eval~2 shifts sharply under p1 on ARGA and CP in both domains, and on RTC in Telecom. Mode stability is weaker if that annotator is taken as the human target. Retail Eval~3 is identical across p0, p1, and p2. It gives no evidence either way on mode stability in that domain.

\subsection{Significant Divergence in Safety Metrics}
The largest human-judge gaps are on Safety Recall (SR) and Irreversible Action Safety (IAS). In Telecom the human scores are several times the judge scores (Table~\ref{tab:human_llm_ratio_telecom}). Humans record far more safety events than either judge. Our companion benchmark MM-$\tau$-p\textsuperscript{2}~\cite{b9} helps explain why. Both GPT-4.1 and GPT-5 produced low and inconsistent safety precision and recall. This happened on identical conversations under identical rubric prompts. It was clearest in escalation cases, such as SIM-lock cases in telecom. There a transfer to a human is the safe and correct action, not an agent failure. The rubric text for a necessary escalation is hard to state without ambiguity. The judges then swing between two readings of the same behavior. One reading credits the agent for handing a high-risk action to a human. The other penalizes the agent for not finishing the task itself. This inconsistency suppresses the recall of real safety events. A human annotator can use situational judgment instead. Safety-critical evaluation should therefore keep human oversight even inside an automated pipeline.

The annotator tables show that this high human baseline is not the work of one scorer. In Telecom all three annotators give high IAS, from 0.794 to 1.000, and high SR, from 0.779 to 1.043. The spread is wider under p1. Even so, no annotator comes near the judge level implied by Table~\ref{tab:human_llm_ratio_telecom}. Human IAS near 0.8 to 1.0 with ratios of 4 to 6 implies judge scores near 0.15 to 0.25. In Retail the annotators are lower, from 0.620 to 0.818 for IAS and 0.436 to 0.857 for SR. That matches the smaller Retail ratios, which stay below 2.

The Retail case is weaker and should not be overstated. Dividing the Eval~1 score by the Retail ratios puts the judge scores near $0.38$ to $0.54$ for IAS and $0.36$ to $0.49$ for SR. Most annotator values are above that band. One is not. Under p2, Eval~2 gives SR $=0.436$, below the $0.490$ implied for GPT-4.1 in the same cell. In Retail the direction of the safety gap therefore depends on the reference annotator in at least one configuration. Telecom has no such case. Every Telecom annotator value is far above the implied judge range. The case for human oversight on safety rests mainly on Telecom. There the gap is large, it holds for all three annotators, and irreversible actions such as SIM-lock or plan-change escalations are more frequent and more costly.

\subsection{Mixed Leadership Between GPT-4.1 and GPT-5}
Neither GPT-4.1 nor GPT-5 leads the other across the board. Alignment with humans shifts by metric. This follows from how the two models are built. GPT-4.1 is a non-reasoning model tuned for low latency and predictable, direct responses \cite{b11, b13}. It does not plan before answering. It gives a fast and literal reading of the rubric on each turn. That helps it track humans on well-specified metrics. It leaves less room to weigh context on ambiguous cases. GPT-5 is a unified reasoning system with an internal planning stage \cite{b12, b10, b13}. It is trained for agentic work such as multi-step tool calls and task coordination. OpenAI reports that this training also reduced sycophancy and cut overconfident answers on tasks the model cannot complete \cite{b12}.

These design choices show up in the scores. GPT-5 is more willing to credit an agent for reasonable effort before escalating. That lifts its goal-completion scores, such as CFA, closer to human levels. The same behavior hurts it on safety metrics, where it under-flags unconfirmed irreversible actions. GPT-4.1 is more conservative and shows the opposite pattern. It tracks humans better on some safety-adjacent judgments. It diverges more on metrics that reward partial progress. The two models are therefore two design points in one family. GPT-4.1 favors speed and predictability \cite{b11}. GPT-5 favors deeper reasoning and agentic judgment \cite{b12}. This trade-off is documented across GPT generations \cite{b10}. It is a more likely cause of the mixed leadership pattern than a simple capability gap.

This comparison uses Eval~1, and the choice of annotator matters in a limited and predictable way. Both judges are scored as $h/a_j$ against the same human value $h$. Replacing Eval~1 with another annotator multiplies both ratios by the same constant. The ordering of the two judges by ratio size therefore cannot change. What can change is which judge is closer to parity. Rescaling can move one ratio across $1$ while the other stays on the same side. This only matters for metrics whose ratios are already near $1$, such as Retail CFA at 0.986 to 1.394. The risk is highest on ARGA, RTC, and CP, where the annotator spread is widest. We did not recompute the per-model ratios under Eval~2 or Eval~3.

Evaluation pipelines may therefore do better with metric-specific or ensemble judging than with one default model. Different metrics capture different parts of conversational performance. Specialized or combined judges can raise reliability and reduce metric-specific bias. This matters most in safety-critical cases.

\subsection{Human Superiority in Recovery Turn Count Estimation}\label{SCM} 
Recovery Turn Count (RTC) is one of the largest differences between human and automated evaluation. Human evaluators assign substantially higher RTC values, while LLM judges often produce values below one. That is hard to reconcile with the definition of recovery, which requires one or more turns after an error.

This gap fits a pattern seen in other work on LLM judges. Agreement with human raters is not uniform. It varies by what is being judged. One large study of empathic communication found expert-LLM agreement from a weighted kappa of 0.17 to 0.86 across 21 dimensions \cite{b16}. Some dimensions were rated almost as well by the LLM as by a second human. Others were rated far worse. RTC looks like one of the harder ones. Scoring it means tracing an error back through several turns. It then means counting forward to the turn where the error was resolved. That is a multi-step tracking task, not a single-turn judgment. Industry evaluations report the same drop-off on harder tasks. Agreement on open-ended, multi-step reasoning falls below half, against over 80 percent on simpler tasks \cite{b17}. A judge scoring a conversation in one pass can lose track of where the error started and where recovery ended. It may count only the turn that contains the fix. A human reviewer can hold the whole exchange in mind and trace the repair more slowly. That is a plausible reason why judge RTC values cluster near zero while human values run higher.

The annotator tables both support and limit that account. All three humans report RTC above 1 in every domain and configuration. The finding that humans count at least one recovery turn therefore does not depend on Eval~1. In Telecom the gap is clear. Dividing the Eval~1 score by the ratios in Table~\ref{tab:human_llm_ratio_telecom} places both judges between 0.22 and 1.22 turns. That is below every annotator value in the matching cell.

Retail is different. The idea of RTC as a uniform human advantage does not survive the annotator data. The implied GPT-5 estimate exceeds at least one annotator in each Retail configuration. The clearest case is p0, where GPT-5 implies about 1.75 turns against 1.385 for Eval~1. In Retail the ordering between human and judge RTC therefore depends on the reference annotator.

Humans also disagree with each other on the count. Telecom p1 is the extreme case, at 1.318, 4.875, and 2.030 turns. Telecom p0 and p2 are much tighter. Retail spreads in every configuration. It is 1.385, 2.500, and 2.381 under p0, and 2.951, 3.618, and 2.381 under p2. RTC has the second widest annotator spread of any metric here. It is hard for human raters as well as for judges. The case for human review of recovery therefore rests on Telecom. It also rests on using more than one rater, not on treating any single human count as correct.

\subsection{Baseline Differences Rather Than Structural Disagreement}\label{sec:baseline_diff}
Human and judge scores often keep the same relative order across metrics even when the absolute values differ. Metrics that score high for humans also tend to score high for the judges. Metrics that are weak stay weak for both. This matches other work on LLM-based rating. A psychometric study compared human, GPT, and Claude raters on large-scale writing assessment \cite{b14}. Generalizability coefficients for relative decisions, such as ranking one essay above another, were consistently higher than reliability coefficients for absolute, fixed-standard scoring. The two rater groups agreed on order but not on scale. The same study found that both human and LLM raters are more reliable at comparative judgments than at criterion-referenced ones \cite{b14}. A separate study using multiple LLM judges reports the same effect \cite{b15}. Rank correlations with expert rankings stayed stable across a wide range of scoring thresholds. Absolute scores moved with the threshold, but the rank order barely did. Rank-order agreement is therefore more stable than absolute-score agreement. The practical task is not to replace human evaluation. It is to calibrate judge scores to a human baseline.

Calibration still needs a stable baseline. The annotator tables show that stability depends on the metric. We use the mean relative range defined in Section~\ref{sec:manual_crosscheck}. CFA ($0.11$), TE ($0.14$), CR ($0.15$), and UES ($0.15$) are the most reproducible across annotators. Eval~1 is a defensible calibration target for those four. ARGA ($0.77$), RTC ($0.46$), and CP ($0.27$) are the least reproducible. A correction factor fit to Eval~1 on those three would shift if Eval~2 were used instead. IAS ($0.15$) and SR ($0.20$) fall in between. Their human score is reasonably reproducible and their human-judge gap is large. That combination suits calibration well.

Two caveats apply. Reproducibility here is measured on metric-level aggregates, not on individual conversations, so it is an upper bound on agreement. Retail Eval~3 gives no independent check at all, since it does not change from p0 to p2. Calibration is a reasonable approach where annotators agree. Where they do not, the rubric given to human raters also needs work. A scale correction cannot fix an unstable target. 

\subsection{Human-Autoeval Score Correlation Analysis}
\label{sec:human_autoeval_correlation}
Table~\ref{tab:backcalc_scores} shows something the ratios alone could not. For most metrics both evaluators land on the same side of the scale, even when the values differ. TE and CR are high for both in both domains. ARGA is low for humans and higher for the judges in both domains. That repeats the pattern in Sections III.A and III.B. SR and IAS behave the same way. They are high for humans and low for both judges. The gap is widest in Telecom, where the judge scores fall below 0.21 and the human scores stay above 0.85. The Human column in that table is the Eval~1 baseline, not the mean of the three annotators.

\begin{table*}[htbp]
\caption{Human and Autoeval Scores, Pooled Across Configurations}
\label{tab:backcalc_scores}
\centering
\scriptsize
\setlength{\tabcolsep}{4pt}
\renewcommand{\arraystretch}{1.2}
\begin{tabular}{|l|c|c|c|c|c|c|}
\hline
\multirow{2}{*}{\textbf{Metric}} & \multicolumn{3}{c|}{\textbf{Telecom}} & \multicolumn{3}{c|}{\textbf{Retail}} \\
\cline{2-7}
 & \textbf{Human} & \textbf{GPT-4.1} & \textbf{GPT-5} & \textbf{Human} & \textbf{GPT-4.1} & \textbf{GPT-5} \\
\hline
CFA & 0.839 & 0.480 & 0.485 & 0.467 & 0.383 & 0.436 \\
\hline
CP  & 0.416 & 0.151 & 0.143 & 0.441 & 0.784 & 0.382 \\
\hline
CR  & 1.000 & 0.931 & 0.995 & 0.838 & 0.967 & 0.990 \\
\hline
IAS & 0.993 & 0.199 & 0.186 & 0.694 & 0.511 & 0.433 \\
\hline
SR  & 0.860 & 0.187 & 0.204 & 0.591 & 0.470 & 0.399 \\
\hline
ARGA & 0.051 & 0.270 & 0.317 & 0.236 & 0.511 & 0.456 \\
\hline
TE  & 0.875 & 0.963 & 0.985 & 0.822 & 0.982 & 0.986 \\
\hline
UES & 0.220 & 0.440 & 0.429 & 0.221 & 0.479 & 0.447 \\
\hline
ERR & 0.283 & 0.263 & 0.137 & 0.267 & 0.054 & 0.075 \\
\hline
RTC & 1.666 & 0.415 & 0.926 & 2.508 & 1.877 & 2.017 \\
\hline
\end{tabular}

\vspace{2pt}
\raggedright

\end{table*}

We next measure how closely the two evaluators track each other in magnitude. We compute the Pearson and Spearman correlations between the human and autoeval score vectors across the ten metrics. We do this twice. The first run uses the configuration-pooled scores in Table~\ref{tab:backcalc_scores}, giving $N = 10$ metrics. The second treats each configuration as a separate observation, giving $N = 30$. The second run therefore captures variation across configurations as well as across metrics. Table~\ref{tab:backcalc_correlation} reports both.

\begin{table}[htbp]
\caption{Human-Autoeval Correlation on Back-Calculated Scores}
\label{tab:backcalc_correlation}
\centering
\scriptsize
\setlength{\tabcolsep}{4pt}
\renewcommand{\arraystretch}{1.2}
\begin{tabular}{|l|l|c|c|c|c|}
\hline
\textbf{Domain} & \textbf{Judge} & \textbf{$r$ (N=10)} & \textbf{$\rho$ (N=10)} & \textbf{$r$ (N=30)} & \textbf{$\rho$ (N=30)} \\
\hline
Telecom & GPT-4.1 & 0.295 & 0.236 & 0.318 & 0.224 \\
\hline
Telecom & GPT-5 & 0.592 & 0.503 & 0.606 & 0.460 \\
\hline
Retail & GPT-4.1 & 0.912 & 0.624 & 0.909 & 0.697 \\
\hline
Retail & GPT-5 & 0.943 & 0.539 & 0.854 & 0.661 \\
\hline
\end{tabular}

\vspace{2pt}
\raggedright
\footnotesize Note: $N=10$ uses configuration-pooled scores (one point per metric). $N=30$ uses each of the three configurations as a separate observation per metric. $r$ = Pearson correlation, $\rho$ = Spearman rank correlation.
\end{table}

The two domains differ sharply here. In Retail the human and autoeval scores are strongly correlated in magnitude for both judges. Pearson $r$ is $0.912$ for GPT-4.1 and $0.943$ for GPT-5, both significant at $p < 0.001$. The two evaluators still disagree on exact values. A metric that scores relatively high for humans also scores relatively high for the judge. The same holds at the low end of the scale. In Telecom the relationship is much weaker. For GPT-4.1 it is not significant, with $r = 0.295$ and $p = 0.408$ at $N=10$, and $r = 0.318$ and $p = 0.087$ at $N=30$. GPT-5 does better in Telecom, with $r = 0.592$ at $N=10$ and $r = 0.606$ at $N=30$. The second of these is significant. Both still fall well short of the Retail values.

This adds to the stability finding in Section~\ref{sec:correlation} rather than contradicting it. That section asked whether a metric keeps its ratio as the configuration changes. It found that most metrics, including SR and IAS, keep the same direction across modes for Eval~1. The correlation here asks a different question. Across the ten metrics, does the judge score rise and fall with the human score? A metric can hold a stable ratio across configurations while the full set of ten still fails to line up between evaluators. That is what happens in Telecom. The Telecom judges are not scaling every metric by a roughly constant factor. GPT-4.1 in particular differs from humans on which metrics count as strong and which count as weak. This fits the domain comparison in Section III.C, where Telecom showed the larger safety divergence. The correlation shows that the disagreement is not limited to SR and IAS. It extends to how well the whole metric profile matches.

The weak Telecom correlation is unlikely to be an Eval~1 artifact. IAS and SR are the main Telecom mismatch in Table~\ref{tab:backcalc_scores}. Both are high for all three annotators in every configuration. Using Eval~2 or Eval~3 would not pull those human points down to the judge range near 0.19. It would move ARGA, RTC, and CP, which are the least stable human points in the correlation. We did not recompute Table~\ref{tab:backcalc_correlation} under those alternative baselines.

This adds a domain caveat to the calibration argument in Section~\ref{sec:baseline_diff}. Consider a linear calibration of the form $\hat{h}_i = \alpha_m \cdot a_i + \beta_m$, fit separately for each metric $m$. That remains reasonable for Retail. The strong linear relationship there suggests the judge captures the right signal and mainly needs rescaling. Telecom is harder. The correlation is weaker, and for GPT-4.1 it is not significant. A per-metric linear correction may not close the gap on its own. Calibration in Telecom may also need rubric-level fixes for RTC and CP. For GPT-4.1 the relationship across metrics is closer to noise than to a linear trend. Any such fit should also treat the human target as uncertain on RTC, ARGA, and CP. Those three have the widest annotator spread in Tables~\ref{tab:manual_p0}, \ref{tab:manual_p1}, and \ref{tab:manual_p2}.

\subsection{ASR Error Analysis }
\begin{itemize}
\item Significant ASR-related issues were observed in recognizing customer names, order IDs, phone numbers, email IDs, and URLs across telecom and retail conversations.
\item In telecom logs, users providing phone numbers in the requested xxx-xxx-xxxx format were frequently misrecognized, resulting in repeated prompts and increased transfers to human agents.
\item Spacing-related transcription errors were observed, affecting the accurate interpretation of user inputs.  
\item User email IDs were frequently misrecognized, leading to failures in information capture and task completion.
\item In some cases, the system was unable to recover even when users followed instructions clearly, such as spelling out names to improve recognition accuracy.
\end{itemize}
ASR errors fell mainly on alphanumeric content. This covers names, numbers, URLs, and email addresses. These failures hurt task completion, raised recovery attempts, and caused avoidable handoffs to human agents. Better ASR accuracy on structured inputs is therefore important for voice-agent reliability.

The annotator scores agree with this picture, with one caveat. ARGA measures goal achievement after an ASR error. It is low for every annotator in Telecom, from 0.037 to 0.273. Retail ARGA is higher but stays at or below 0.342 in every cell except Eval~2 under p1, which gives 0.652. RTC counts the turns spent recovering. It is above 1 for every annotator, domain, and configuration. All three annotators therefore agree on two things. ASR errors are often not recovered, and recovery takes more than one turn. They do not agree on the size of either effect. ARGA and RTC have the two widest annotator spreads in this study. Neither should be treated as a single ground-truth number without a second scoring pass. 

\section{Implications}
LLM-as-Judge systems can already support large-scale evaluation workflows. This matches current industry practice. A recent survey found that 92 percent of teams run LLM judges inside their continuous integration and deployment pipelines \cite{b17}. Teams that use them well report over twice the reliability of teams that avoid them, and they catch more incidents rather than fewer \cite{b17}. Academic work agrees. A large psychometric study of human, GPT, and Claude raters found acceptable reliability for relative decisions with as few as one or two raters per group \cite{b14}. Our results point the same way. Most metrics in Table~\ref{tab:human_llm_ratio_telecom} hold a stable ratio between human and judge scores across modes. Calibration needs exactly that. A stable ratio means a judge score can be mapped to an expected human score with a simple correction. The results therefore support automated evaluation at scale for most metrics. Scale alone does not close every gap.

Six considerations follow.

\begin{itemize}
\item \textbf{Safety metrics need human validation}: SR and IAS diverge most from human judgment. All three annotators score them high in Telecom, so this is not one rater's view.
\item \textbf{Recovery metrics need better modelling}: RTC is unreliable in current judges. It is also among the least reproducible metrics for human raters, second only to ARGA. It needs a clearer rubric on both sides.
\item \textbf{Calibration can improve alignment}: Metric trends stay stable across evaluators. A per-metric correction factor can therefore map judge scores toward human-equivalent scores.
\item \textbf{Calibration works best where the human baseline is stable}: A correction fit on CFA, TE, CR, or UES is robust to the choice of annotator. One fit on ARGA, RTC, or CP is not, because the human target itself moves.
\item \textbf{Calibration is also domain-dependent}: Section~\ref{sec:human_autoeval_correlation} shows a simple per-metric linear calibration is well supported in Retail, where $r > 0.9$ for both judges. It is weaker in Telecom, and for GPT-4.1 it is not significant. Rescaling may be enough in some domains. Others may need rubric revisions as well.
\item \textbf{No single judge model is best}: GPT-4.1 and GPT-5 trade places by metric. GPT-5 tracks humans more closely on goal-completion metrics such as CFA. GPT-4.1 tracks humans more closely on some safety-adjacent judgments. Metric-specific or ensemble judging may therefore beat a single default model.
\end{itemize}
LLM evaluators are best used as scalable first-pass tools. Human review still matters for safety-sensitive work and for recovery analysis. Calibration and judge choice should be set per domain and per metric rather than applied uniformly.

\section{Conclusion}

This study compared human evaluators with GPT-4.1 and GPT-5 on conversational voice agents. Across p0, p1, and p2, both LLM judges held stable metric-level trends. Their relative assessments often followed human judgments. Their absolute scores did not.

The largest human-judge differences were on Safety Recall, Irreversible Action Safety, and Recovery Turn Count. The judges recorded far fewer safety events than human raters. They also counted fewer recovery turns. Neither model led across the full metric set. GPT-4.1 reads the rubric quickly and literally. That helps on well-specified metrics. GPT-5 reasons further and credits partial agent effort. That helps on goal completion but under-flags unconfirmed safety risks. Evaluator quality therefore depends on the metric. No single judge model is uniformly better.

Scoring the same conversations with three annotators shows how far these conclusions can be pushed. The safety result is the most robust. All three annotators scored IAS and SR high in Telecom. Every value is well above the level implied for either judge. That gap does not depend on the annotator. The recovery result is weaker than it first looked. Every annotator counted more than one recovery turn. In Retail, though, the implied GPT-5 estimate exceeded at least one annotator in each configuration. The direction of the Retail RTC gap therefore depends on the reference annotator.

The annotator tables also separate two things that are easy to confuse. Human-judge divergence is largest on IAS and SR. Disagreement between human raters is largest on ARGA, RTC, and CP. A metric can be scored consistently by humans and still be scored very differently by a judge. The reverse also happens. Calibration is well supported on CFA, TE, CR, and UES, where annotators agree closely. It is weaker on ARGA, RTC, and CP, where the human target itself moves. For those three the human rubric needs work, not only the judge score.

These results support LLM judges for scalable, first-pass evaluation. Human review is still needed for safety-critical decisions and for recovery analysis in Telecom. A hybrid pipeline remains the practical choice. Metric-specific calibration can improve alignment. It is more straightforward in Retail than in Telecom, where the cross-metric correlation is weaker.

Two limits should be kept in view. Agreement here was measured on metric-level aggregates, not on individual conversations, so it is an upper bound. Retail Eval~3 is identical across the three configurations and gives no independent check in that domain. Future work should measure conversation-level agreement with a standard reliability coefficient. It should recompute the ratios under each annotator. It should also tighten the rubrics for ambiguous safety and recovery cases.

\section{Future Work}
The current benchmark does not model temporal and interaction-level phenomena unique to voice, such as missed response windows, prolonged silence, interruptions, overtalk, overlapping speech, and unsuccessful barge-in handling. These are hard to capture from text transcripts alone, since their evaluation depends on acoustic and timing information, yet they can affect recovery behavior and user experience and may drive repeated requests or call abandonment. Future work will extend the framework to audio signals, speaker-turn boundaries, and timestamp data, enabling metrics for response-window adherence, interruption detection, overtalk frequency, and barge-in success, and will separate failures caused by ASR from those caused by dialogue-management or response-generation policies.

We also plan to test whether audio-aware or multi-modal LLM judges assess these voice-specific behaviors more reliably than text-only judges, and to validate the proposed calibration framework across larger datasets, additional tasks, and more diverse speakers and acoustic conditions. Finally, future work should refine rubrics for safety-sensitive escalation and multi-turn recovery to determine which voice-specific metrics can be automated reliably and which still require human oversight.

\section{Acknowledgment}
Authors acknowledge Peeyush Aggarwal for his support towards and Aditya Choudhary for data analysis related to LLM based evaluation.

\appendices
\section{Evaluation Metric Definitions}
\label{app:metric_defs}
\begin{itemize}
    \item \textbf{Critical Field Accuracy (CFA)}: Accuracy on error-sensitive entities (e.g., order ID, phone number, plan identifier) whose incorrect capture can invalidate task success.
    \item \textbf{Confirmation Precision (CP)}: The fraction of requested clarifications/confirmations that were actually necessary; a low CP indicates the agent over-clarifies when context was already sufficient.
    \item \textbf{Confirmation Recall (CR)}: The fraction of required clarifications/confirmations that were actually requested by the agent; a low CR means the agent proceeded on ambiguous input without asking.
    \item \textbf{Safety Recall (SR)}: Consistency with which the agent requests confirmation when required (e.g., under low ASR confidence or ambiguous intent), computed as confirmations requested over confirmation-required cases.
    \item \textbf{Irreversible Action Safety (IAS)}: The proportion of high-risk, irreversible actions (cancellations, charges, plan changes) that were executed only after explicit user confirmation; an IAS below 1.0 flags a critical safety failure.
    \item \textbf{Recovery Turn Count (RTC)}: The average number of conversational turns needed to recover from an error, covering ASR misrecognitions, tool failures, and incorrect agent actions.
    \item \textbf{Turn Efficiency (TE)}: The ratio of the optimal number of turns to the actual number of turns taken to complete a task, with values closer to 1.0 indicating efficient resolution.
    \item \textbf{Turn Overhead (TO)}: Extra turns incurred in voice versus text interactions, computed as $\frac{T_{\text{voice}} - T_{\text{text}}}{T_{\text{text}}}$. TO $< 0.2$ is minimal; TO $> 0.5$ indicates excessive voice friction.
    \item \textbf{User Experience Score (UES)}: A count of user repetitions, corrections, or restatements (e.g., re-spelling a name); high UES signals poor user experience even when the task ultimately succeeds.
    \item \textbf{ASR-Robust Goal Achievement (ARGA)}: The probability of achieving the task goal given that an ASR error occurred, isolating the agent's recovery capability from raw transcription accuracy.
    \item \textbf{Modality Robustness Score (MRS)}: Degradation from text to voice, computed as $\frac{\text{Pass}^k_{\text{voice}}}{\text{Pass}^k_{\text{text}}}$. MRS $= 1.0$ indicates no degradation; MRS $< 0.7$ suggests the agent is not voice-ready.
    \item \textbf{Error Recovery Rate (ERR)}: The proportion of all detected errors, across ASR misrecognitions, tool failures, and wrong agent actions, that were successfully recovered via clarification, retry, or undo.
\end{itemize}


\begin{thebibliography}{00}
\bibitem{b1} S. E. Finch and J. D. Choi, Towards Unified Dialogue System Evaluation: A Comprehensive Analysis of Current Evaluation Protocols, SIGDIAL, 2020. ACL Anthology
\bibitem{b2} T. Ji et al., Achieving Reliable Human Assessment of Open-Domain Dialogue Systems, ACL, 2022. ACL Anthology
\bibitem{b3} L. Zheng et al., Judging LLM-as-a-Judge with MT-Bench and Chatbot Arena, 2023. Paper
\bibitem{b4} J. Gu et al., A Survey on LLM-as-a-Judge, 2024. Paper 
\bibitem{b5} B. Pang et al., Towards Holistic and Automatic Evaluation of Open-Domain Dialogue Generation, ACL, 2020. ACL Anthology
\bibitem{b6} L. Shi et al., Judging the Judges: A Systematic Study of Position Bias in LLM-as-a-Judge, 2024. Paper
\bibitem{b7} H. Wei et al., Systematic Evaluation of LLM-as-a-Judge in LLM Alignment Tasks: Explainable Metrics and Diverse Prompt Templates, 2024. Paper
\bibitem{b8} S. Guan et al., Evaluating LLM-based Agents for Multi-Turn Conversations: A Survey, 2025. Paper
\bibitem{b9} A. Purwar and A. Choudhary, MM-$\tau$-p\textsuperscript{2}: Persona-Adaptive Prompting for Robust Multi-Modal Agent Evaluation in Dual-Control Settings, arXiv:2603.09643, 2026.
\bibitem{b10} H. Afridi, H. Ullah, S. D. Khan, and M. Ullah, From GPT-3 to GPT-5: Mapping their Capabilities, Scope, Limitations, and Consequences, arXiv:2604.10332, 2026.
\bibitem{b11} OpenAI, GPT-4.1 Model Documentation, OpenAI API Docs. \url{https://developers.openai.com/api/docs/models/gpt-4.1}. Accessed: 2026-08-17.
\bibitem{b12} OpenAI, Introducing GPT-5, 2025. \url{https://openai.com/index/introducing-gpt-5/}. Accessed: 2026-08-17.
\bibitem{b13} Microsoft, GPT-5 vs GPT-4.1: Choosing the Right Model for Your Use Case, Microsoft Foundry Docs. \url{https://learn.microsoft.com/en-us/azure/foundry/foundry-models/how-to/model-choice-guide}. Accessed: 2026-08-17.
\bibitem{b14} Y. Wang, J. Huang, L. Du, Y. Guo, Y. Liu, and R. Wang, Evaluating Large Language Models as Raters in Large-Scale Writing Assessments: A Psychometric Framework for Reliability and Validity, Computers and Education: Artificial Intelligence, vol. 9, 100481, 2025.
\bibitem{b15} B. Braun and M. Forell, (Towards) Scalable Reliable Automated Evaluation with Large Language Models, arXiv:2607.28282, 2026.
\bibitem{b16} J. P. Kim, Augmenting Human Evaluation with LLM Judges: How Many Human Reviews Do You Need?, arXiv:2605.16354, 2026.
\bibitem{b17} P. Bhavsar, LLM-as-a-Judge vs Human Evaluation: When to Use Each (And Why Elite Teams Use Both), Galileo AI Blog, 2026. \url{https://galileo.ai/blog/llm-as-a-judge-vs-human-evaluation}. Accessed: 2026-08-18.
\bibitem{b18} A. Purwar and A. Choudhary,  i-LAVA: Insights on Low Latency Voice-2-Voice Architecture for Agents, arxiv:2509.20971, 2025.
\bibitem{b19} A. Purwar and A. Choudhary, FOCAL: A Novel Benchmarking Technique for
Multi-modal Agents, arxiv:2601.07367, 2026
\end{thebibliography}
\end{document}